\documentclass[conference]{IEEEtran}

\usepackage{copyright}

\usepackage[english]{babel}
\usepackage{amssymb}
\usepackage{amsmath}
\usepackage{hyperref}
\usepackage{graphicx}
\usepackage{latexsym}
\usepackage{algorithm}
\usepackage{algpseudocode}
\usepackage{todonotes}
\usepackage{newclude}
\usepackage{multicol}
\usepackage{tabularx}
\usepackage{booktabs}
\usepackage{threeparttable}

\usepackage{enumitem}
\usepackage{listings}
\usepackage{combelow}
\usepackage{subfigure}
\usepackage{bigfoot}
\usepackage{multirow}
    
\usepackage[acronym]{glossaries}
\newacronym{vo}{VO}{Visual Odometry}
\newacronym{adas}{ADAS}{Advanced Driver Assistance Systems}
\newacronym{lidar}{LiDAR}{Light Detection and Ranging}
\newacronym{orbi}{ORBI}{Occluded Road Boundary Inference}
\newacronym{dnn}{DNN}{Deep Neural Network}
\newacronym{gnss}{GNSS}{Global Navigation Satellite Systems}
\newacronym{ldw}{LDW}{Lane Departure Warning}
\newacronym{lka}{LKA}{Lane Keeping Assist}
\newacronym{gps}{GPS}{Global Positioning System}
\newacronym{slam}{SLAM}{Simultaneous Localisation and Mapping}
\newacronym{ro}{RO}{Radar Odometry}
\newacronym{lo}{LO}{Laser Odometry}
\newacronym{vrbd}{VRBD}{Visible Road Boundary Detection}
\newacronym{roi}{ROI}{region of interest}
\newacronym{icp}{ICP}{Iterative Closest Point}
\newacronym{ipm}{IPM}{Inverse Perspective Mapping}
\newacronym{6dof}{6DoF}{six degree-of-freedom}

\usepackage[binary-units]{siunitx}
\usepackage{cleveref}
\crefname{table}{Table}{Tables}
\crefname{figure}{Figure}{Figures}
\crefname{section}{Section}{Sections}

\usepackage{xcolor}

\newcommand{\firstdspair}{{\sc Experiment 1}}
\newcommand{\seconddspair}{{\sc Experiment 2}}

\begin{document}

%------------------------------------------------------------------
\title{LiDAR Lateral Localisation \\ Despite Challenging Occlusion from Traffic}
\author{Tarlan Suleymanov, Matthew Gadd, Lars Kunze, and Paul Newman\\
Oxford Robotics Institute, Dept. Engineering Science, University of Oxford, UK.\\\texttt{\{tarlan,mattgadd,lars,pnewman\}@robots.ox.ac.uk}}
\maketitle
%------------------------------------------------------------------

\copyrightnotice

%------------------------------------------------------------------
\begin{abstract}
This paper presents a system for improving the robustness of \acrshort{lidar} lateral localisation systems.
This is made possible by including detections of road boundaries which are invisible to the sensor (due to occlusion, e.g. traffic) but can be located by our \acrlong{orbi} \acrlong{dnn}.
We show an example application in which fusion of a camera stream is used to initialise the lateral localisation.
We demonstrate over four driven forays through central Oxford -- totalling \SI{40}{\kilo\metre} of driving -- a gain in performance that inferring of occluded road boundaries brings.
\end{abstract}
%------------------------------------------------------------------

%------------------------------------------------------------------
%------------------------------------------------------------------

%------------------------------------------------------------------
\section{Introduction}
%------------------------------------------------------------------

Autonomous vehicles are required to perceive their surrounding environment and know their location in the world before they can plan a path to safely navigate to a desired location.

Indeed, accurate lateral localisation is crucial for many \gls{adas} such as \gls{ldw}, \gls{lka}, and Parking Assist systems.
However, these systems are susceptible to cluttered environments, where traffic and other obstructions are likely to disrupt the robust performance required for safe operation of the vehicle.

\begin{figure}
    \centering
    \includegraphics[width=\columnwidth]{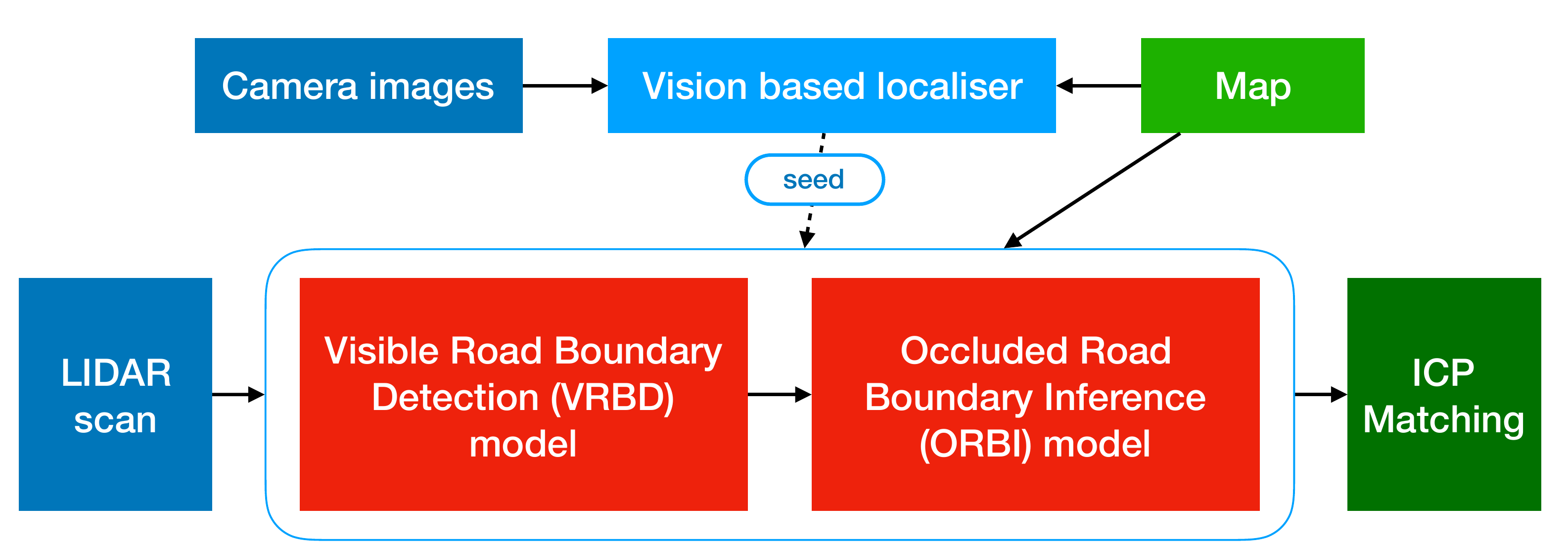}
    \caption{On overview of the pipeline proposed in this paper.
    After localisation is coarsely initialised, the live \gls{lidar} scan is passed through our \gls{vrbd} and \gls{orbi} models.
    This gives us not only visible road boundaries but also inferred locations for the occluded parts of those road boundaries.
    These are then matched to a map which has been similarly processed for visible and occluded road boundaries.
    Note that localisation is coarsely initialised by a camera stream but this part of the system is interchangeable with e.g. \gls{lidar} place recognition systems such as~\cite{kim20191}.}
    \label{fig:localisation_pipeline}
\end{figure}

\begin{figure}
    \centering
    \includegraphics[width=\columnwidth]{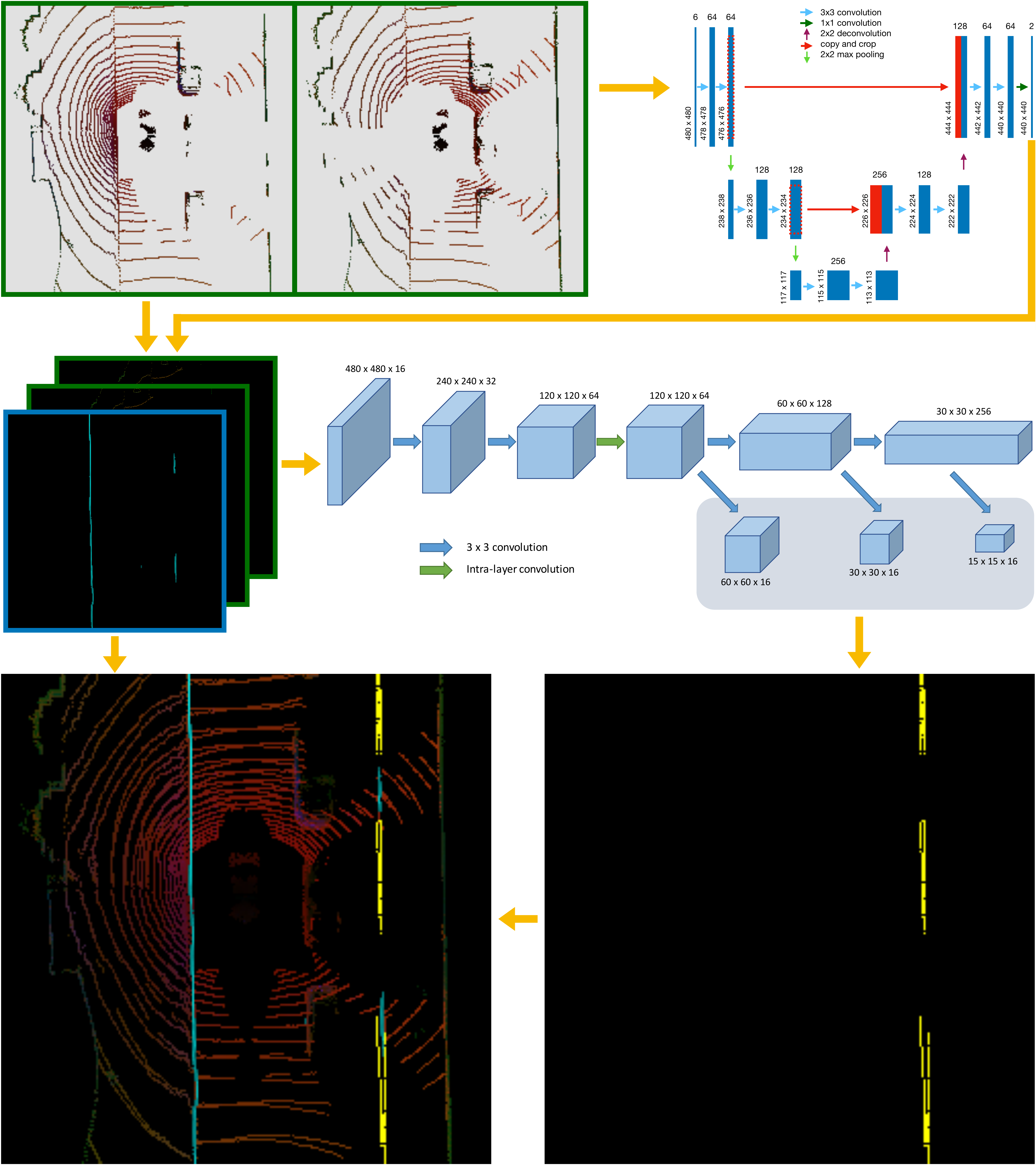}
    \caption{Our LIDAR-based coupled approach for road boundary detection. Given a pair of \gls{ipm}  images from left and right LIDARs, the fully convolutional VRBD model detects visible road boundaries and then passes to the ORBI model for inference of occluded road boundaries. The second model contains 3 base layers, intra-layer convolutions and 3 layers of parameterised multi-scale predictions at the end.}
    \label{fig:road_boundary_detection_pipeline}
\end{figure}

To this end, in previous work, we presented road segmentation~\cite{SuleymanovIROS2016}, road boundary detection~\cite{Suleymanov2018Curb,Suleymanov2019Curb}, and scene understanding~\cite{2018ITSC_kunze}, which were about perceiving the environment of autonomous vehicles.
Now, we take this perceived information -- in this case detected road boundaries -- and apply it to solve one of the fundamental tasks of autonomous driving and \gls{adas}: localisation, or more specifically, lateral localisation.
\cref{fig:localisation_pipeline} shows an overview of our pipeline, while~\cref{fig:road_boundary_detection_pipeline} shows an overview of the \gls{vrbd} and \gls{orbi} models that enable this localisation system.

This paper proceeds by reviewing related literature in~\cref{sec:related}.
\cref{sec:method} describes our approach to include invisible or occluded road boundaries in lateral localisation systems.
We describe in~\cref{sec:experimental} details for implementation, evaluation, and our dataset.
\cref{sec:results} discusses results from such an evaluation.
\cref{sec:concl,sec:future} summarise the findings and suggest further avenues for investigation.

%------------------------------------------------------------------
\section{Related Work}
\label{sec:related}
%------------------------------------------------------------------

Mapping and localisation techniques can be classified into two groups: global and relative~\cite{Thrun2001RobustMonteCarlo}.

\begin{figure}
    \centering
    \includegraphics[width=\columnwidth]{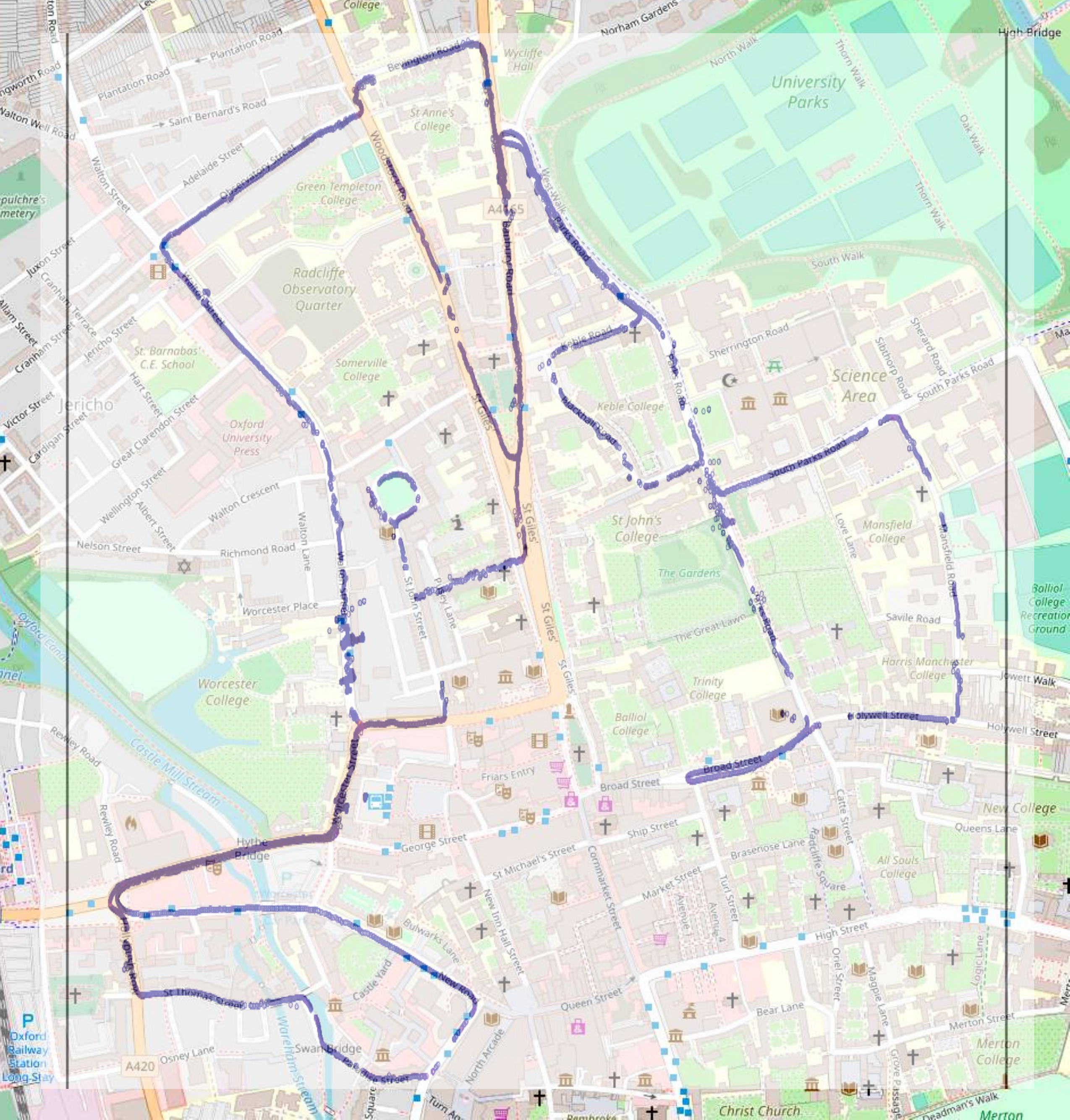}
    \caption{Samples of vehicle position are overlaid on a digital map using \gls{gps} coordinates.
    Although the dataset is a complete loop, we observe many gaps along the route.}
    \label{fig:localisation_vis_occ_all_map}
\end{figure}

\emph{Global techniques} for localisation of a robot can be achieved using \gls{gnss}, but it is not precise enough for autonomous driving as the accuracy is worse than 2-3 metres in an open-sky environment~\cite{Ghallabi2018LiDARBasedLaneMarking}.
Indeed, consider~\cref{fig:localisation_vis_occ_all_map}.
We observe that there are many gaps along the route when samples from the \textit{Oxford RobotCar Dataset}~\cite{RobotCarDatasetIJRR} are overlaid on a digital map using \gls{gps} coordinates.
\emph{Relative techniques} based on \gls{slam}~\cite{grisetti2010tutorial} or ego-motion estimation~\cite{borenstein1996measurement} do not employ a global frame of reference but are less susceptible to drift and scale better with vast environments.

Depending on the type of input sensor these approaches can be further categorised by their use of cameras (passive sensors) and \glspl{lidar} or radars (active sensors).
\emph{Camera-based techniques}, such as \gls{vo}~\cite{nister2004visual}, are sensitive to lighting conditions, shadows, illumination, under- and overexposure~\cite{mcmanus2014shady}.
Using dense maps in \emph{\gls{lidar}-based approaches}, such as \gls{lo}~\cite{zhang2014loam}, is usually very accurate but computationally expensive for running in real-time.
There is burgeoning interest in \emph{radar-based techniques}, including \gls{ro}~\cite{cen2018precise,2019ITSC_aldera,2019ICRA_aldera} and localisation~\cite{KidnappedRadarArXiv}.
As of yet, these techniques do not offer a rich enough understanding of scene semantics to be employed for road boundary problems.

Finally, regardless of the sensor used, \emph{feature matching} is often used to match inputs from the sensors to maps.
Indeed, Lane markings, traffic signs, feature points or road boundaries can be used as features for localisation~\cite{sun20193d}.
The long and continuous shape of road boundaries makes them stable and robust features for localisation in the lateral direction as they capture the structure of roads.

In this paper we use sparse \gls{lidar}-based detected road boundaries for computational ease and infer occluded road boundaries to meet the challenge of partially or fully occlusion by obstacles or other road users.

%------------------------------------------------------------------
\section{Methodology}
\label{sec:method}
%------------------------------------------------------------------

\begin{figure}
    \centering
    \includegraphics[width=\columnwidth]{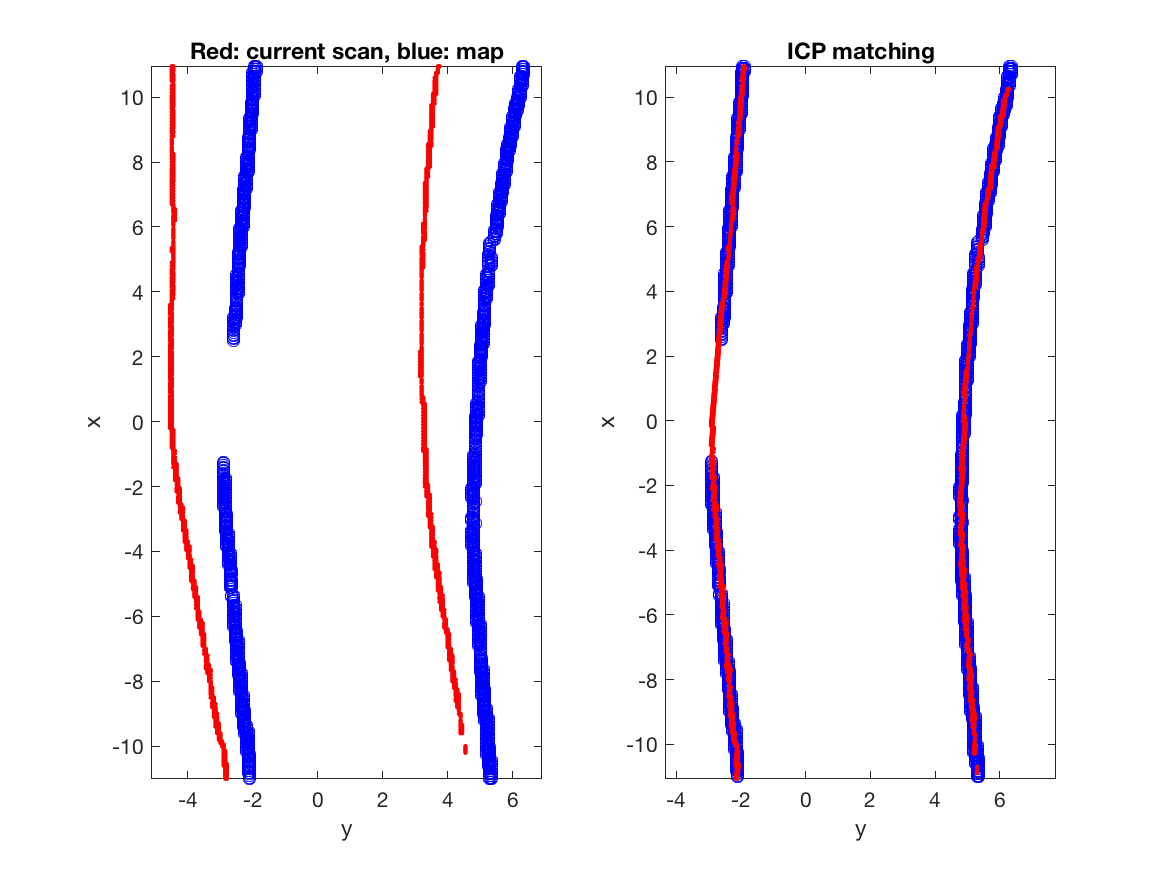}
    \caption{Given the detected road boundaries mask of a scan, it is binarised and transformed into a point cloud to match with a scan from the map.
    \gls{icp} is used to match the road boundaries and estimate the transformations between the frames.
    This example shows the accurate estimation of the transformation despite the undetected section of the road boundaries in the map frame.}
    \label{fig:icp_matching}
\end{figure}

Our cross-track localisation approach, the workflow for which is shown in~\cref{fig:localisation_pipeline}, is designed to:

\begin{enumerate}
    \item Demonstrate usability of the outputs of a \gls{lidar}-based road boundary detection approach for lateral localisation, and
    \item Demonstrate a gain in performance that inferring of occluded road boundaries could bring.
\end{enumerate}

This section details (in~\cref{sec:method:models,sec:method:matching}) the two subsystems inherent to our pipeline.
As to be detailed in~\cref{sec:experimental:curation}, the system acts between a map and a live trajectory, each taken from a different dataset (or collection of sensory records captured during a foray by the vehicle).
We begin by describing a preprocessing step for all map and live \gls{lidar} scans.

\subsection{\gls{lidar} \gls{ipm}}

Labels and 3D \gls{lidar} scans are transformed into 2D bird’s eye view images -- or \glspl{ipm} -- to obtain input images and their road boundary masks.
Note that the 3D \gls{lidar} scans are trimmed to keep only the points that are close to the road surface before transforming them into \gls{ipm} images.
Our test vehicle has two 3D \glspl{lidar} and we generate \gls{ipm} images for both of them before combining them in a single frame, accounting for the \gls{6dof} extrinsic offset in sensor mount positions.

\begin{figure}
    \centering
    \includegraphics[width=0.49\columnwidth]{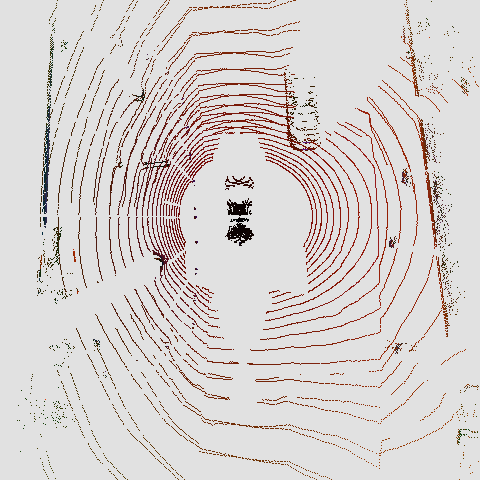}
    \includegraphics[width=0.49\columnwidth]{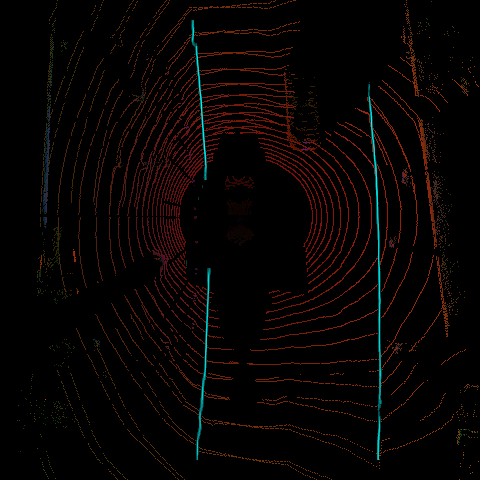}
    \includegraphics[width=0.49\columnwidth]{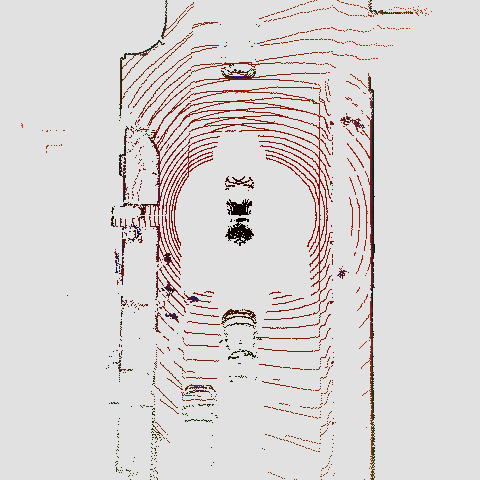}
    \includegraphics[width=0.49\columnwidth]{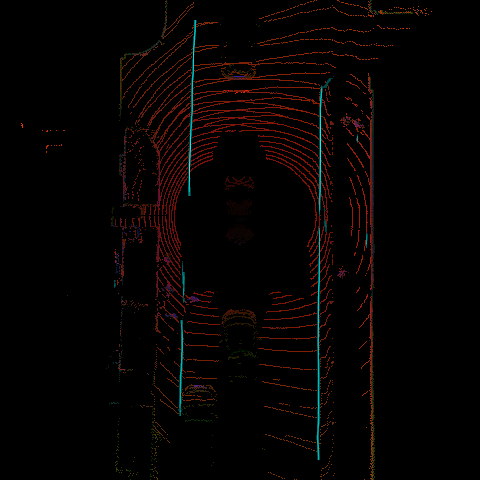}
    \caption{Output samples of detected visible road boundaries by the VRBD model with a ROI of 24x24 squared metre.}
    \label{fig:vis_only_road_boundary_detection}
\end{figure}

\begin{figure}
    \centering
    \includegraphics[width=0.49\columnwidth]{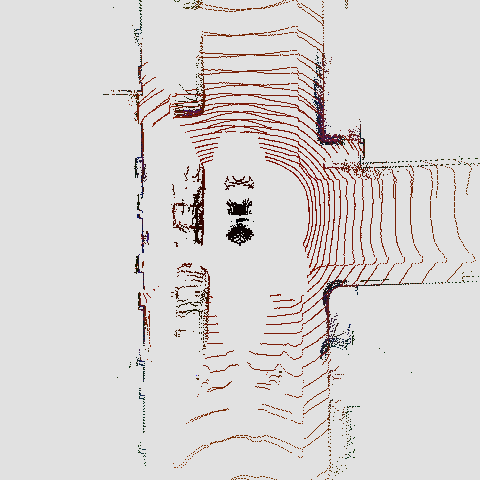}
    \includegraphics[width=0.49\columnwidth]{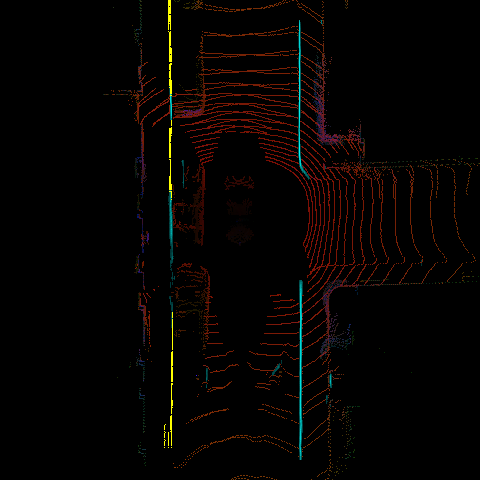}
    \includegraphics[width=0.49\columnwidth]{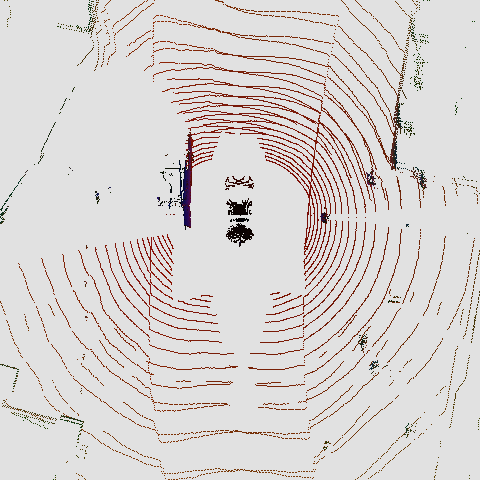}
    \includegraphics[width=0.49\columnwidth]{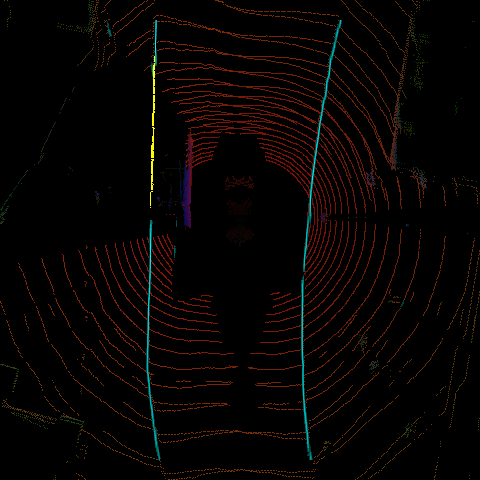}
    \caption{Output samples of detected visible and inferred occluded road boundaries by the VRBD and ORBI models with a ROI of 24x24 squared metre.}
    \label{fig:vis_and_occ_road_boundary_detection}
\end{figure}

\subsection{Visible and occluded road boundary detection}
\label{sec:method:models}

We adopt our \gls{lidar}-based \gls{vrbd} and \gls{orbi} models presented in~\cite{Suleymanov2018Curb,Suleymanov2019Curb} to run the inference over the map dataset with the \gls{roi} of 24x24 squared metre, where 1 px correspond to \SI{5}{\centi\metre} in the real world.
Outputs of the detected road boundaries are stored in the map with corresponding timestamps.
\cref{fig:road_boundary_detection_pipeline} shows and describes this in more detail.
\cref{fig:vis_only_road_boundary_detection,fig:vis_and_occ_road_boundary_detection} show examples of visible-only road boundary detection as well as inference of the position of occluded road boundaries.

\subsection{Map matching}
\label{sec:method:matching}

The second dataset is used as a live input and the detected road boundaries from the second dataset are matched against the map dataset to perform lateral localisation.
We assume that the initial guess of the location of the vehicle in the map are provided by a vision-based localiser~\cite{LinegarICRA2015}.
Note that the vision-based localiser only provides timestamps of corresponding images from the map without providing initial pose.
We use the \gls{icp} algorithm~\cite{Besl1992ICP} to perform the matching process and estimate the transformations between the live inputs and the map.
\gls{icp} is a well-known algorithm that is used for matching point clouds, where the algorithm iteratively updates the transformation between two point clouds to minimise the distance between them.
We adopt \gls{icp} to estimate the transformation between live and map samples.
The detected road boundary masks are binarised and converted into point clouds and then matched with \gls{icp} as shown in~\cref{fig:icp_matching}.

%------------------------------------------------------------------
\section{Experimental Setup}
\label{sec:experimental}
%------------------------------------------------------------------

This section details our experimental design in obtaining the results to follow in~\cref{sec:results}.

\subsection{Dataset curation}
\label{sec:experimental:curation}

To run our experiments we use a pair of datasets -- one dataset as a map and the other as a live input.
Our experiments cover two pairs of datasets in central Oxford, one taken from~\cite{RobotCarDatasetIJRR} and the other from~\cite{RadarRobotCarDatasetArXiv}.
The pair taken from~\cite{RobotCarDatasetIJRR} are collected in April, 2018 and January 2019, respectively, and we refer to this experiment as~\firstdspair.
The pair taken from~\cite{RadarRobotCarDatasetArXiv} are both collected in January, 2019 and we refer to this experiment as~\seconddspair.

\subsection{Data annotation}

Fine-grained hand annotation of road boundaries from images would be a very time consuming process and it would be impossible to exactly annotate position of occluded road boundaries.
To avoid that we annotate 3D point clouds that are collected using 2D laser.
We then accumulate subsequent vertical laser scans in a coherent coordinate frame using \gls{vo} to estimate vehicle’s motion in order to compute transformations between subsequent scans.

\subsection{Performance metrics}

Our results are relative measures of improved \gls{lidar} lateral localisation when including detections of occluded road boundaries over systems which do not have access to those detections and only use visible road boundaries.
We compare the agreement of the returned poses to the pose available from a state-of-the-art vision-based localiser~\cite{LinegarICRA2015}.

%------------------------------------------------------------------
\section{Results}
\label{sec:results}
%------------------------------------------------------------------

This section presents instrumentation of the metrics discussed in~\cref{sec:experimental} over both experiments: \firstdspair~and~\seconddspair.

\begin{figure}
    \centering
    \includegraphics[width=0.49\columnwidth]{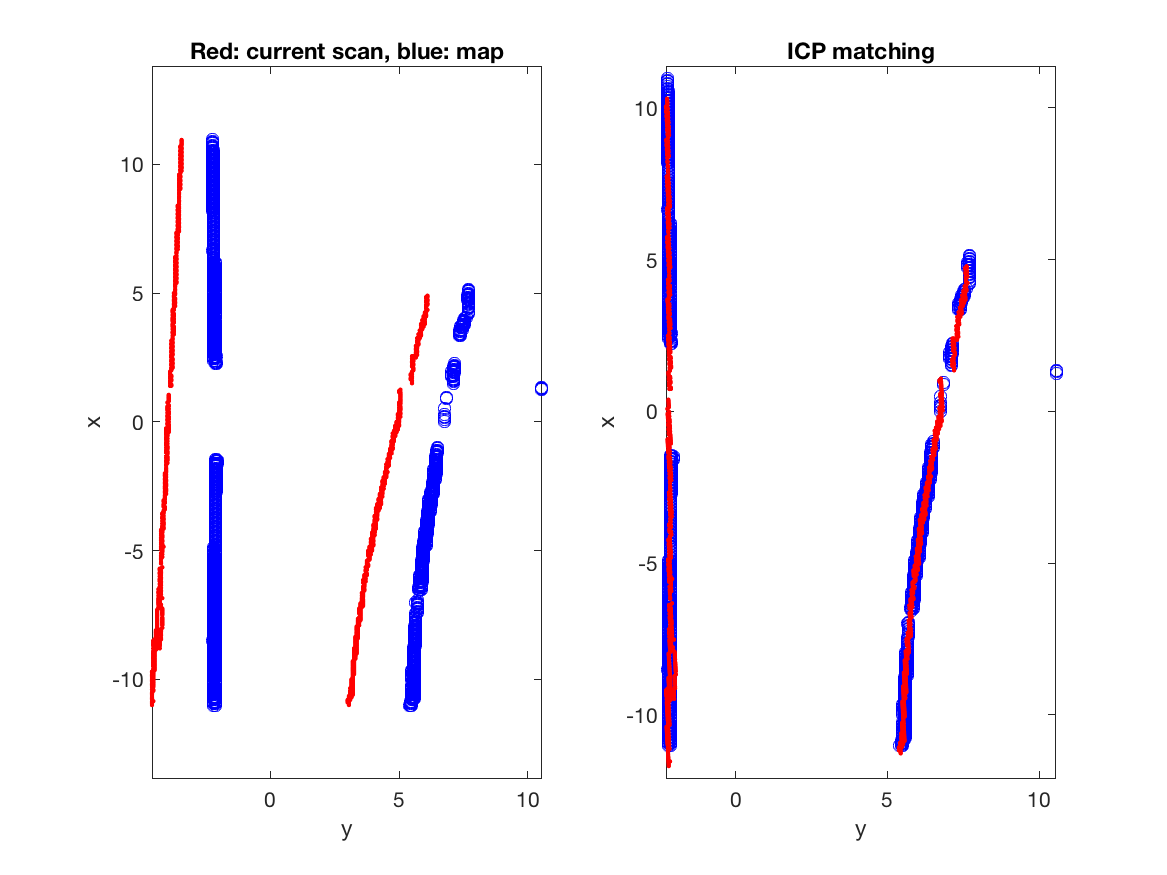}
    \includegraphics[width=0.49\columnwidth]{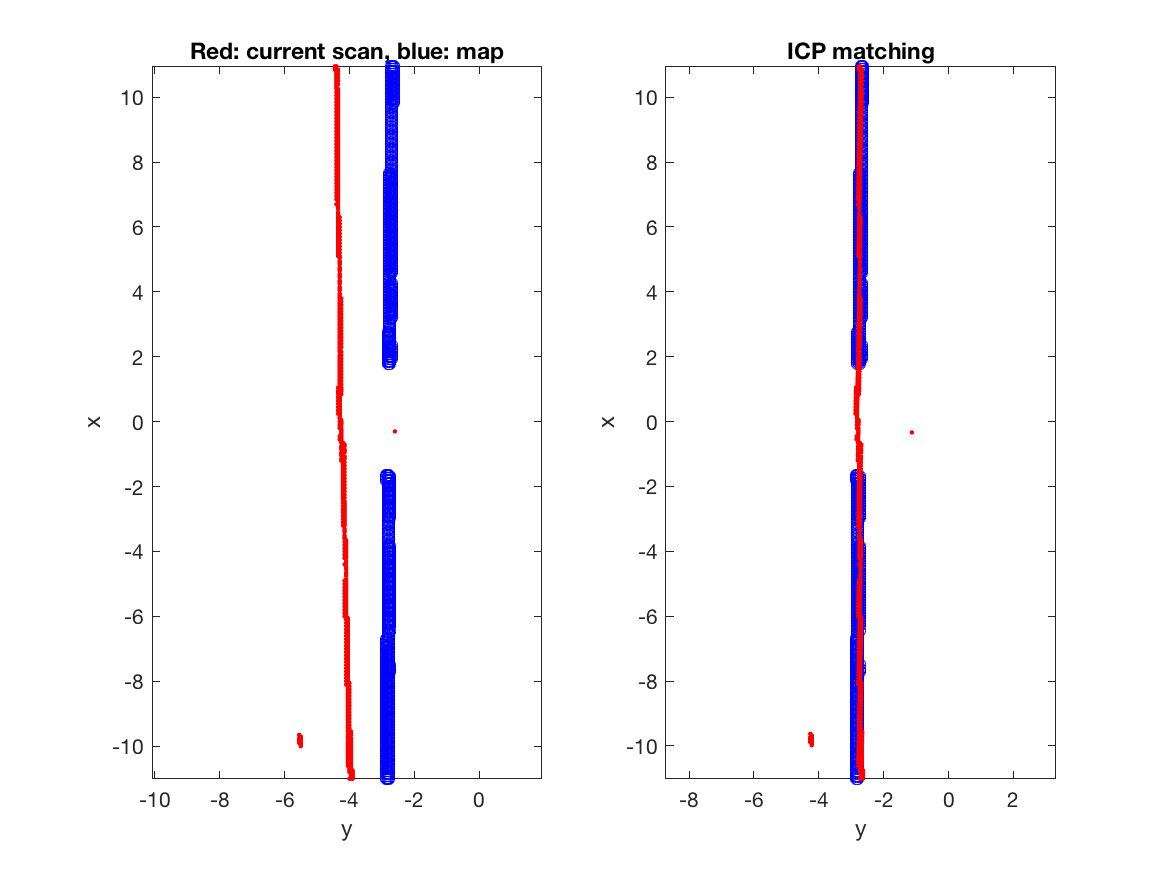}
    \includegraphics[width=0.49\columnwidth]{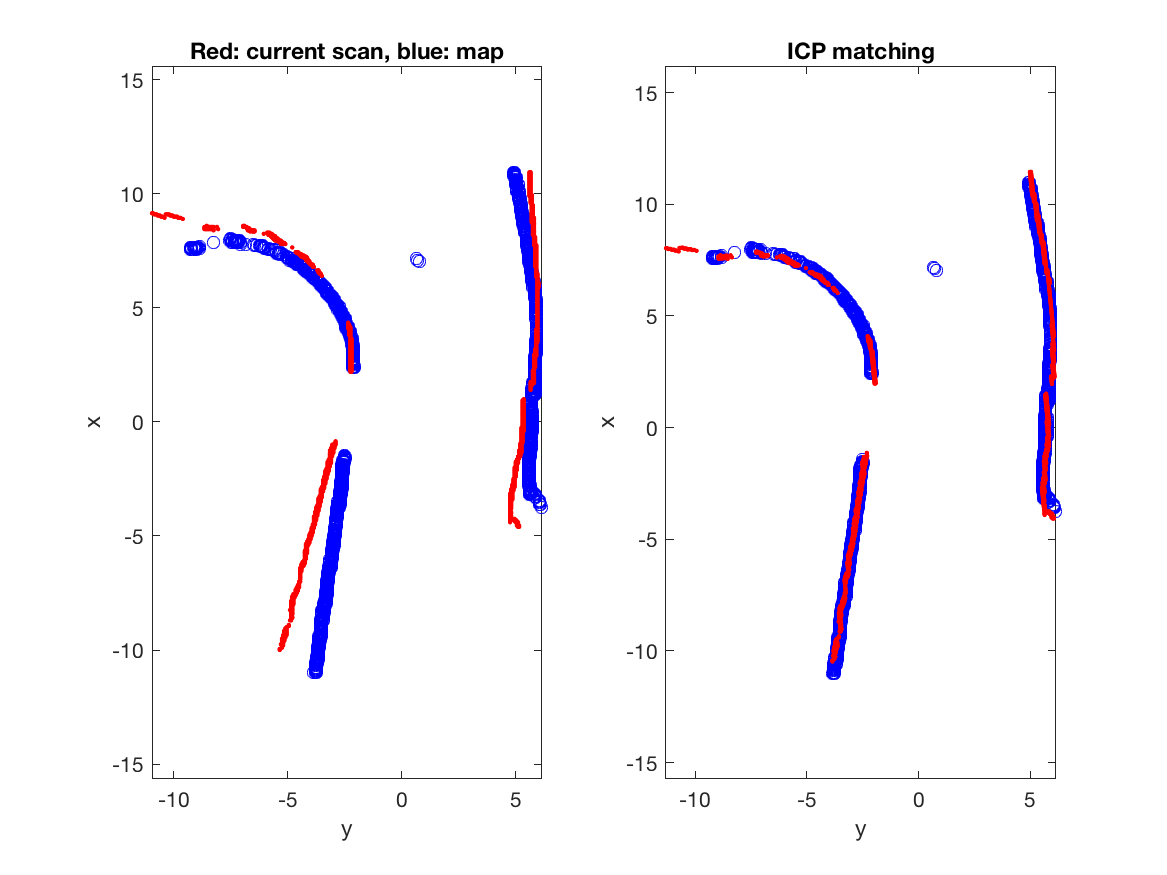}
    \includegraphics[width=0.49\columnwidth]{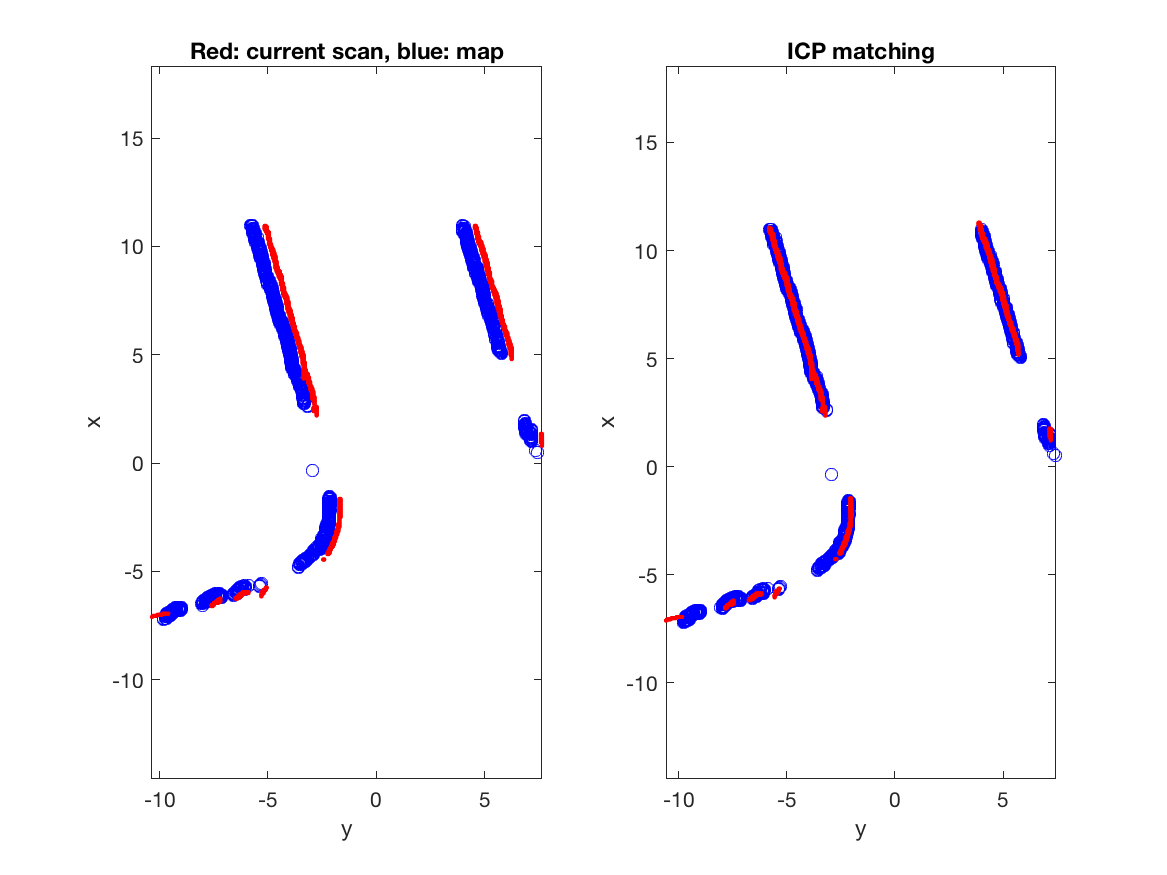}
    \caption{Examples of road boundary based \gls{icp} matching for localisation. These examples demonstrate that \gls{icp} accurately estimates the transformations between samples irrespective of the structure of the detected road boundaries since the detected road boundaries between samples are balanced over the sections of the true boundaries.
    Results obtained by processing the \firstdspair~dataset pair.}
    \label{fig:icp_matching_success}
\end{figure}

We present in~\cref{fig:icp_matching_success} qualitative examples of road boundary based \gls{icp} matching for localisation.
These examples demonstrate that \gls{icp} accurately estimates the transformations between samples irrespective of the structure of the detected road boundaries.
Small amounts of noise in the detection does not change the overall estimation of the transformations given that the road boundaries are detected in a balanced way over the sections of the true boundaries.
\gls{icp} can accurately estimate the transformation in the presence of detected road boundaries on one side of the road.

\begin{figure}
    \centering
    \includegraphics[width=0.49\textwidth]{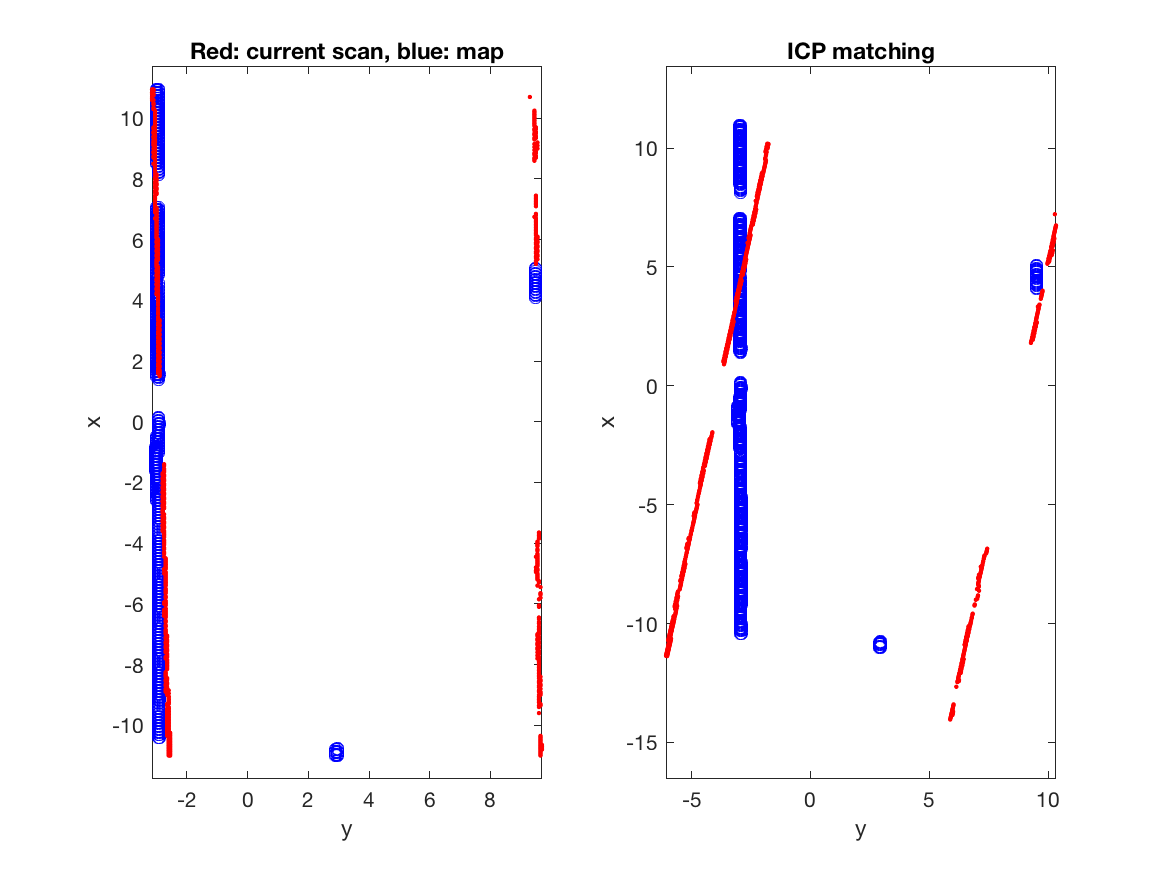}
    \includegraphics[width=0.49\textwidth]{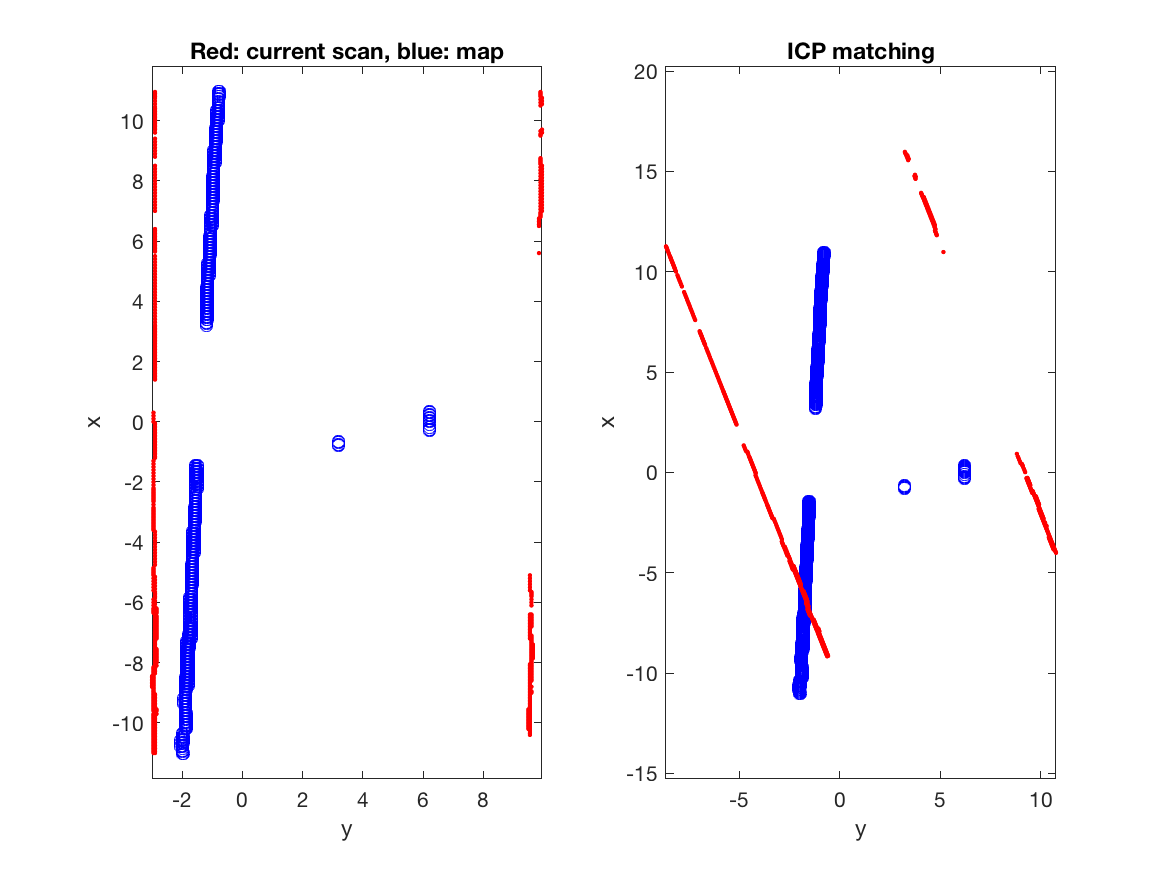}
    \caption{ICP matching failure examples, where the detected road boundaries on the right hand side of the road are unbalanced between samples. Results obtained by processing the \firstdspair~dataset pair.}
    \label{fig:icp_matching_failure}
\end{figure}

\begin{figure}
    \centering
    \includegraphics[width=0.49\textwidth]{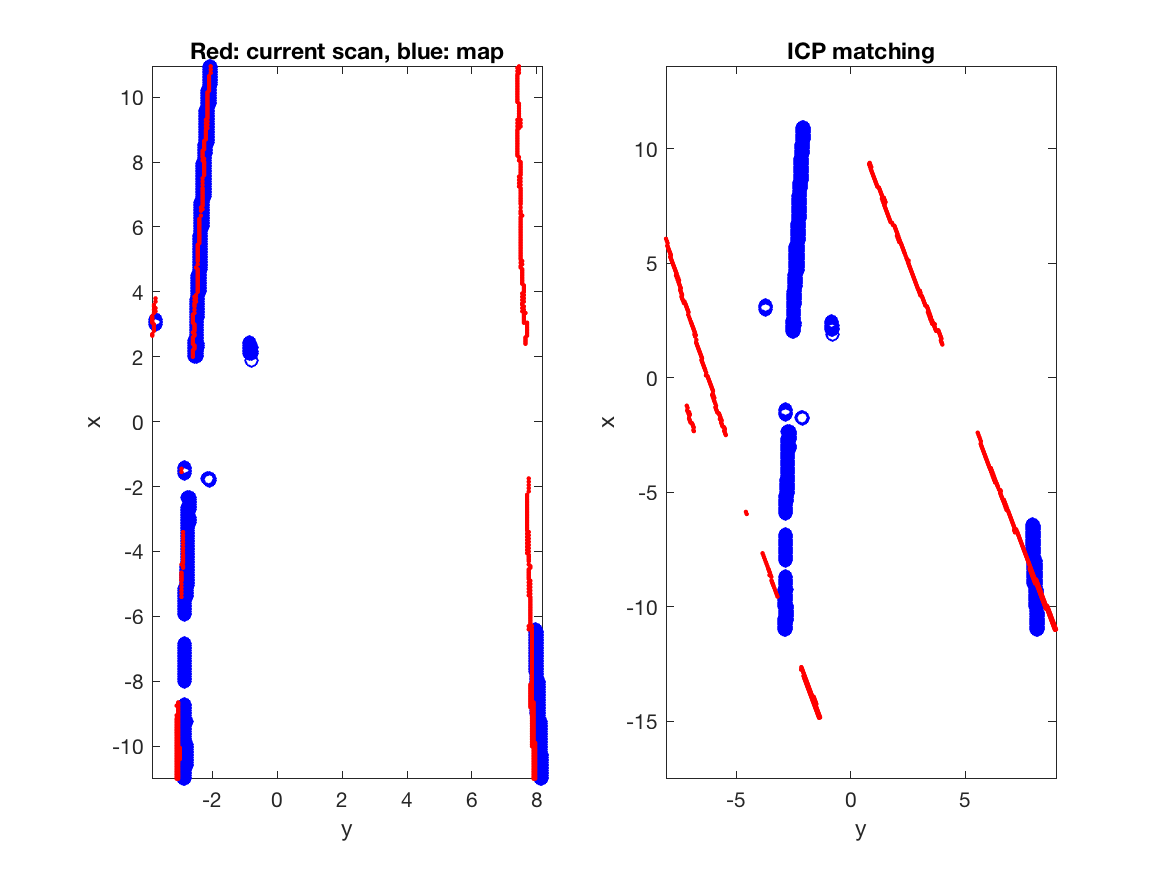}
    \includegraphics[width=0.49\textwidth]{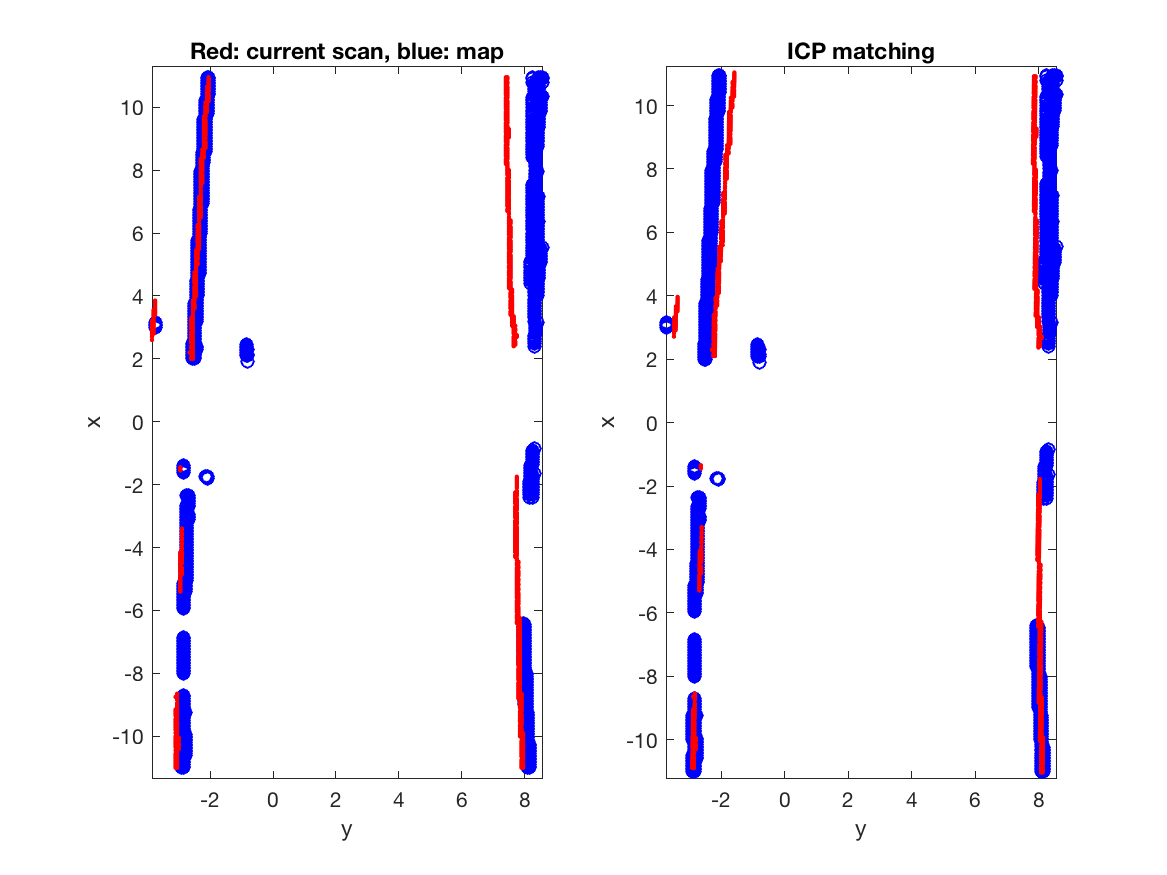}
    \caption{Comparison: \gls{icp} matching based on detected visible road boundaries only (top) and based on all road boundaries (bottom).
    Results obtained by processing the \firstdspair~dataset pair.}
    \label{fig:icp_matching_vis_vs_vis_occ}
\end{figure}

Consider~\cref{fig:icp_matching_failure}.
In these illustrated cases \gls{icp} fails to match the road boundaries accurately to generate transformations between samples. This happens when the detected road boundaries between live and map inputs are unbalanced, which forces \gls{icp} to rotate the live inputs as keeping them parallel is more costly.
This can be fixed using more sophisticated matching techniques.
Consider, however,~\cref{fig:icp_matching_vis_vs_vis_occ} in which we include in the matching detections of occluded road boundaries as proposed in this paper.
Without more sophisticated matching techniques, we are now able to obtain sane transformations.

\begin{table}
\centering
\caption[]{Road boundary based lateral localisation results, comparing localisation based on visible road boundaries only with localisation based on a combination of visible and occluded road boundaries. The results show that using the inferred occluded road boundaries always improves performance. Results obtained by processing the \firstdspair~dataset pair.}
\resizebox{\columnwidth}{!}{
\begin{tabular}{|l|cc|cc|}
\toprule
 & \multicolumn{2}{c|}{Visible only} & \multicolumn{2}{c|}{Visible and occluded} \\
\multirow{-2}{*}{Error range} & Number of samples & Percentage & Number of samples & Percentage \\
\midrule
Within 0.1 metre & 20389 & 65.56\% & 21934 & 70.53\% \\
Within 0.3 metre & 27479 & 88.36\% & 28349 & 91.15\% \\
Within 0.5 metre & 29005 & 93.26\% & 29617 & 95.23\% \\
Within 1 metre & 30124 & 96.86\% & 30504 & 98.08\% \\
\bottomrule
\end{tabular}
\label{tab:localisation_error_distance}}
\end{table}

\begin{table}
\centering
\caption[]{Average lateral localisation error for both the \firstdspair~and~\seconddspair~dataset pairs.}
\resizebox{\columnwidth}{!}{
\begin{tabular}{|l|c|c|c|c|}
\toprule
Matching method & Map dataset & Live dataset & Visible only & Visible and occluded \\
\midrule
ICP & 2018-04-30 & 2019-01-18 & 18.95 cm & 14.72 cm \\
ICP & 2019-01-10-11 & 2019-01-10-12 & 18.54 cm & 12.18 cm \\
ICP with worst rejection & 2019-01-10-11 & 2019-01-10-12 & 9.33 cm & 7.51 cm \\
\bottomrule
\end{tabular}
\label{tab:localisation_error_distance_2}}
\end{table}

\begin{figure}
    \centering
    \includegraphics[width=\columnwidth]{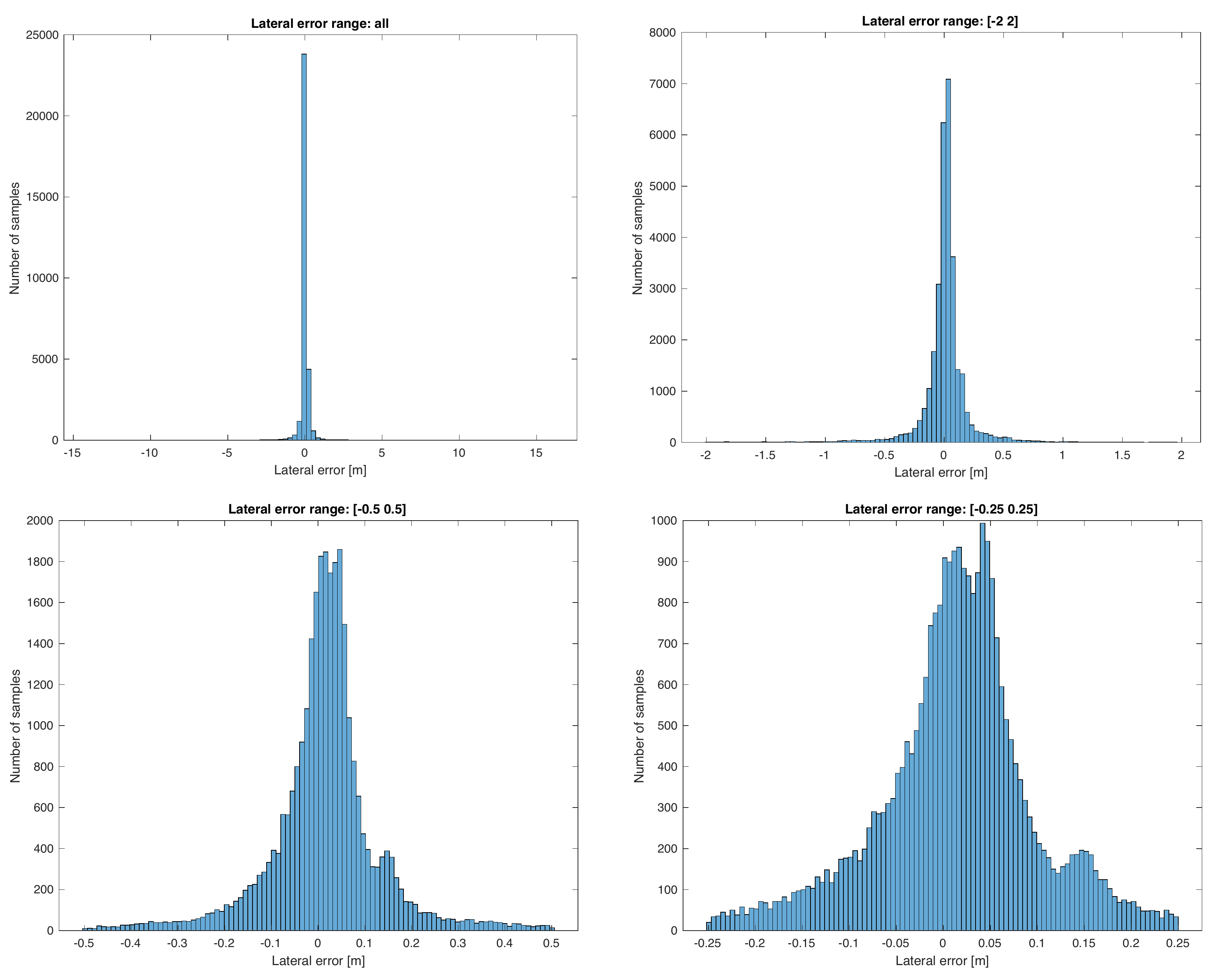}
    \caption{Histograms of road boundary based lateral error. The top left histogram includes all samples from the dataset, while the remaining histograms progressively narrow the displayed error range (horizontal axis). We observe that the majority of the samples (70.53\%) have a maximum lateral error smaller than 10 cm. Results obtained by processing the \firstdspair~dataset pair.}
    \label{fig:localisation_lateral_error}
\end{figure}

\begin{figure}
    \centering
    \includegraphics[width=\columnwidth]{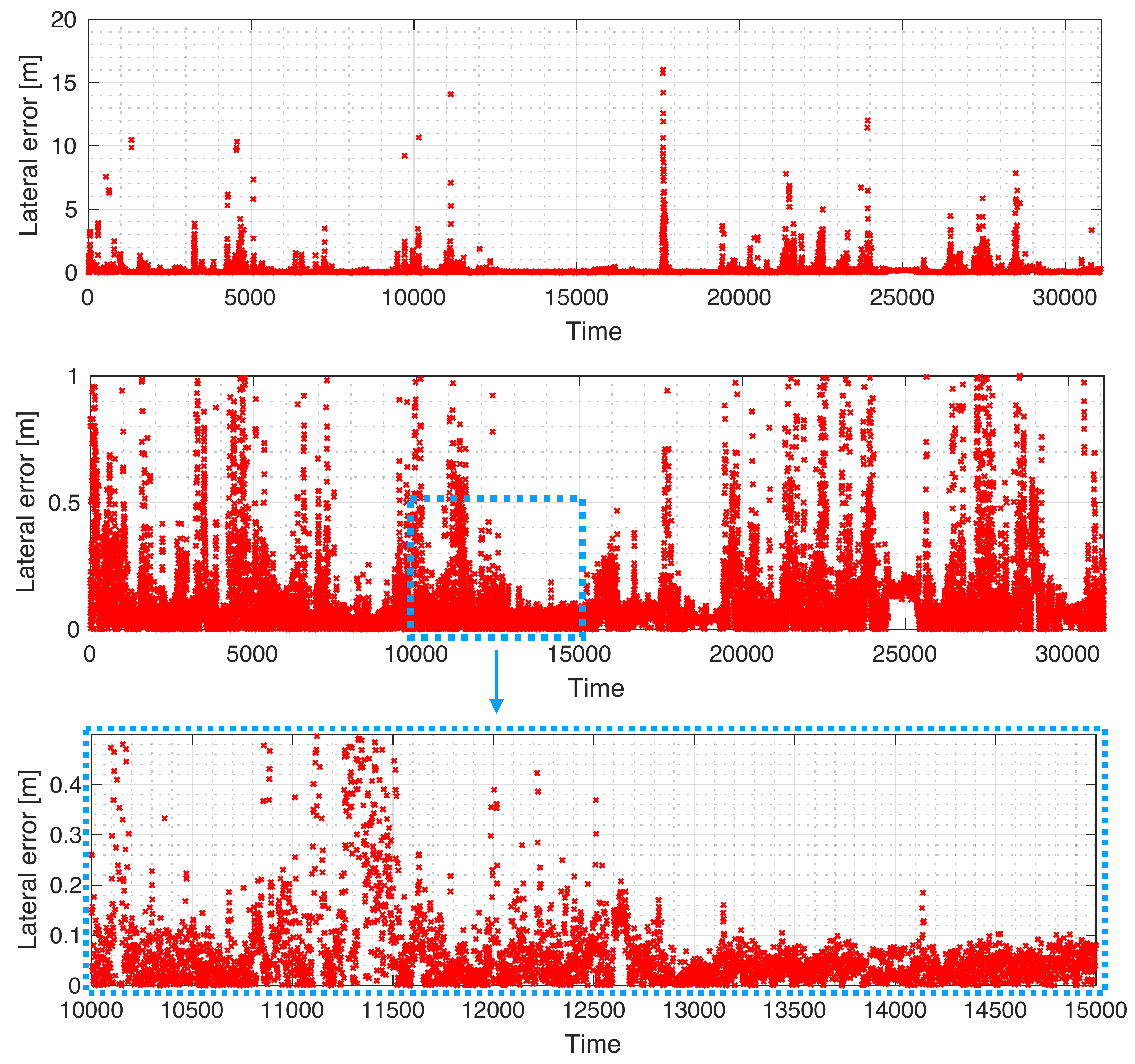}
    \caption{Lateral error of all samples displayed in a timeline (top), where we observe that there are only a small number of peaks where localisation failed. Progressively narrowing the displayed error range (vertical axis) shows that the majority of the samples (70.53\%) have a maximum lateral error within 10 cm. Results obtained by processing the \firstdspair~dataset pair.}
    \label{fig:localisation_timelines}
\end{figure}

\begin{figure*}
    \centering
    \includegraphics[width=\textwidth]{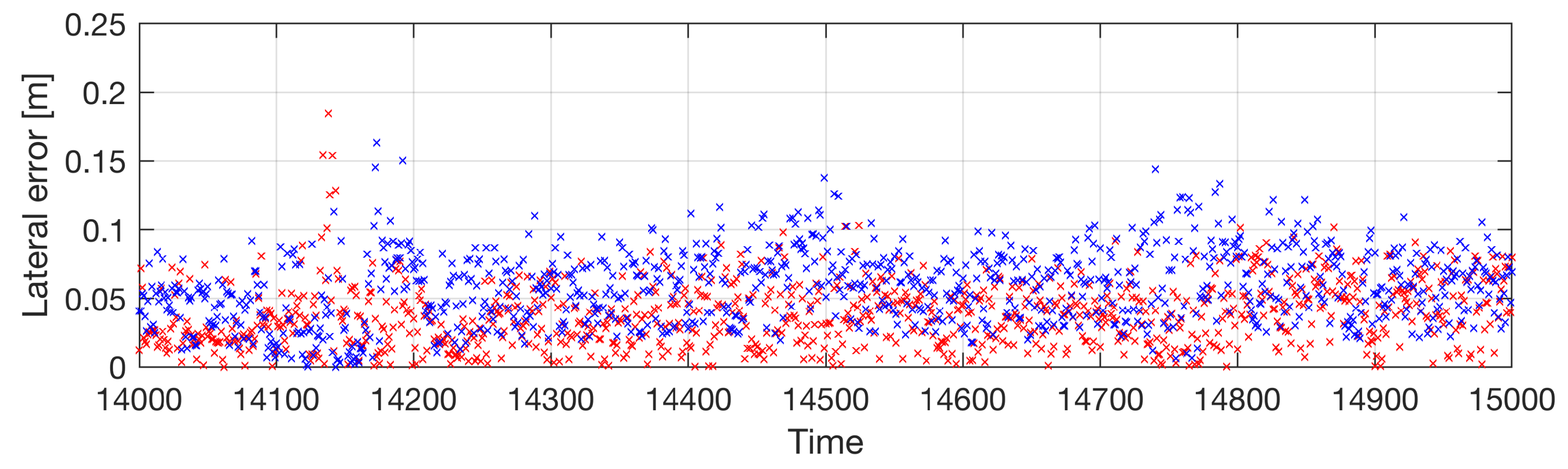}
    \caption{A timeline of 1000 samples, displaying lateral errors based on only visible road boundaries (blue points) and  errors based on all road boundaries (red points). We observe that the blue points are generally larger than the red ones, indicating that using all road boundaries is better for performance. Results obtained by processing the \firstdspair~dataset pair.}
    \label{fig:localisation_vis_vs_vis_occ_timeline}
\end{figure*}

\begin{figure*}
    \centering
    \includegraphics[width=\textwidth]{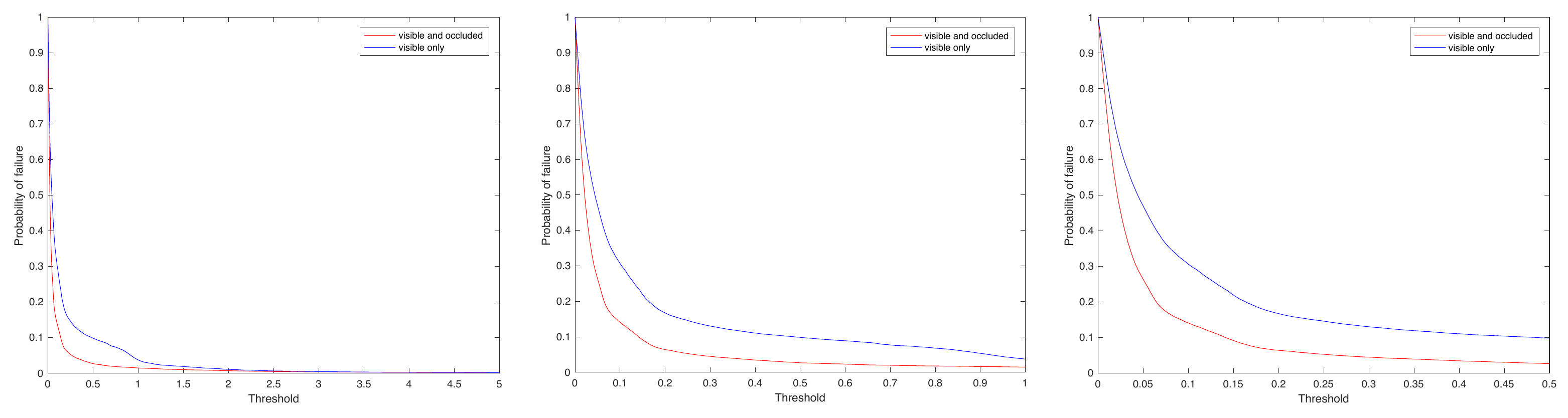}
    \caption{Probability that a localisation failure (measured as a lateral deviation in the $y$-direction) will occur on a given localisation run.
    Results obtained by processing the \seconddspair~dataset pair.}
    \label{fig:probability_of_failure}
\end{figure*}

We compare in~\cref{tab:localisation_error_distance,tab:localisation_error_distance_2,fig:localisation_lateral_error,fig:localisation_timelines,fig:localisation_vis_vs_vis_occ_timeline,fig:probability_of_failure} the estimated road boundary based lateral localisation results with the lateral localisation of the vision-based localiser.
Note that the \gls{ipm} images that are used for estimating the lateral localisation are interpolated to match the timestamps of the camera images that the vision-based localiser uses.

Here, we calculate average lateral error (mean absolute error) and yaw error for the lateral localisation based on:
\begin{enumerate}
    \item Visible road boundaries only, and
    \item A combination of all road boundaries.
\end{enumerate}

For \firstdspair, our results show that the average lateral localisation error based on visible road boundaries only is \SI{18.95}{\centi\metre}.
Including the inferred occluded road boundaries decreases the error by \SI{4.23}{\centi\metre} to \SI{14.72}{\centi\metre}.
Similarly, the yaw error decreases from \SI{0.0332}{\radian} to \SI{0.0206}{\radian}.
Similar boons to performance for \seconddspair~can be read from~\cref{tab:localisation_error_distance_2}.

We analyse in~\cref{fig:localisation_lateral_error,tab:localisation_error_distance} the output results by counting the number of samples that have a lateral error within 1, 0.5, 0.3, and 0.1 metres.
Using the inferred occluded road boundaries increases the percentage of the number of samples within 1 metre from 96.86\% to 98.08\%
Similar gains are achieved for the number of samples within 0.5, 0.3, and 0.1 metres.
Overall, 70.53\% of samples have lateral error less than 10 cm in contrast with the vision-based localiser.
To visualise the lateral error of the samples we will plot them as histograms.

The lateral error of all samples are displayed in the timeline of~\cref{fig:localisation_timelines} to show that the majority of the samples (98.08\%) has an error less than one metre and there are only a small number of peaks where localisation fails.
We also plot a zoomed in section of the timeline to show that the large number of samples (70.53\%) has a lateral error less than 10 cm.

Another timeline plotted in~\cref{fig:localisation_vis_vs_vis_occ_timeline} contains only 1000 samples and shows the importance of occluded road boundaries.
The blue points in the figure represent the lateral error based on only visible road boundaries, where the red points are the localisation errors with all road boundaries.
Overall, we observe that the blue points (only visible road boundaries) have higher values than the red ones (visible and occluded road boundaries).

Finally,~\cref{fig:probability_of_failure} shows distributions of the magnitude of localisation failure for \seconddspair.
These can be understood as the probability that a localisation failure (measured as a lateral deviation) will occur on a given localisation run.
We observe for all ranges of failures the inclusion of inferred positions of occluded road boundaries is beneficial to the robustness of a lateral localisation system.

%------------------------------------------------------------------
\section{Conclusions}
\label{sec:concl}
%------------------------------------------------------------------

In this paper we presented a road boundary based lateral localisation method that provides accurate results using \gls{lidar} data.
The proposed system leverages new methods for detecting occluded road boundaries in sparse \gls{lidar} data to improve \gls{lidar}-based lateral localisation beyond that which is currently possible by na\"{i}ve visible road boundary detection based methods.
This is particularly evident and relevant in environments which are challenged by dense traffic conditions.
Additionally, the proposed system is innovative in comparison to systems which may rely on \gls{gps} initialisation of the lateral localisation system by fusing the initial pose hint from onboard sensors such as cameras (although the method is agnostic to how this hint is sourced), which is particularly important in \gls{gps}-denied environments.
This is also beneficial in comparison to existing systems which perform along-path localisation with \glspl{lidar} and rely in that sense on the detection of objects in the scene (e.g. traffic signs), which may also be occluded (or missing) in particularly challenging environments.
We demonstrated that using inferred occluded road boundaries improves performance in environments which are challenged by the presence of obstructions.

%------------------------------------------------------------------
\section{Future Work}
\label{sec:future}
%------------------------------------------------------------------

In the future we plan to integrate the system on the all-weather platform described in~\cite{kyberd2019} in challenging, unstructured environments, requiring a new labelling regime to identify the boundaries of driveable paths.

%------------------------------------------------------------------
\section*{Acknowledgment}
%------------------------------------------------------------------

Matthew Gadd is supported by Innovate UK under CAV2 -- Stream 1 CRD (DRIVEN).
Lars Kunze is supported by the Assuring Autonomy International Programme, a partnership between Lloyd’s Register Foundation and the University of York.
Paul Newman is supported by UK EPSRC programme grant EP/M019918/1.

%------------------------------------------------------------------
\bibliographystyle{IEEEtran}
\bibliography{biblio}
%------------------------------------------------------------------

\end{document}